\documentclass[letterpaper, conference, 10 pt]{ieeeconf}
\IEEEoverridecommandlockouts
\usepackage{cite}

\usepackage{amsmath,amssymb,amsfonts}
\usepackage{algorithmic}
\usepackage{graphicx}
\usepackage{textcomp}
\usepackage{xcolor}
\usepackage{commath}
\usepackage{multirow}
\usepackage{caption}
\usepackage{subcaption}
\usepackage[linesnumbered,ruled,vlined]{algorithm2e}
\usepackage{optidef}
\let\labelindent\relax

\usepackage{enumitem}
\usepackage{accents}
\usepackage{xurl}

\def\matt#1{\begin{bmatrix}#1\end{bmatrix}}

\def\BibTeX{{\rm B\kern-.05em{\sc i\kern-.025em b}\kern-.08em
		T\kern-.1667em\lower.7ex\hbox{E}\kern-.125emX}}

\title{\LARGE \bf
	DrMaMP: Distributed Real-time Multi-agent Mission Planning in Cluttered Environment
}

\author{Zehui Lu, \ Tianyu Zhou, \ Shaoshuai Mou
	\thanks{The authors are with the School of Aeronautics and Astronautics, Purdue University, IN 47907, USA {\tt\small \{lu846, zhou1043, mous\}@purdue.edu} }
	\thanks{This work is supported in part by NASA University Leadership Initiative (ULI) under grant number 80NSSC20M0161 and funding from Northrop Grumman Corporation. A supplementary video can be found from: {\tt\small{youtu.be/il3YxhXgGac}} Source code can be found from: {\tt\small{github.com/zehuilu/DrMaMP-Distributed-Real-}}\newline {\tt\small{time-Multi-agent-Mission-Planning-Algorithm}}} 
}

\begin{document}

\maketitle
\thispagestyle{empty}
\pagestyle{empty}

\begin{abstract}
	Solving a collision-aware multi-agent mission planning (task allocation and path finding) problem is challenging due to the requirement of real-time computational performance, scalability, and capability of handling static/dynamic obstacles and tasks in a cluttered environment. This paper proposes a distributed real-time (on the order of millisecond) algorithm DrMaMP, which partitions the entire unassigned task set into subsets via approximation and decomposes the original problem into several single-agent mission planning problems. This paper presents experiments with dynamic obstacles and tasks and conducts optimality and scalability comparisons with an existing method, where DrMaMP outperforms the existing method in both indices. Finally, this paper analyzes the computational burden of DrMaMP which is consistent with the observations from comparisons, and presents the optimality gap in small-size problems.
\end{abstract}


\section{Introduction}\label{sec:introduction}

Autonomous unmanned aerial vehicles (UAV) and unmanned ground vehicles (UGV) can replace humans for dangerous tasks such as surveillance and search-and-rescue. Recently, some receding-horizon motion planning methods \cite{herbert2017fastrack, kousik2019safe, tordesillas2019faster} guide an autonomous robot to explore and go to a destination in a complex environment. These methods require a planning hierarchy. On top of this hierarchy, a path planner such as \cite{dijkstra1959note, nash2010lazy} generates a sequence of sparse way-points based on the perception of the environment. Then a motion planner returns collision-free and dynamically feasible trajectories based on the sparse way-points. And the robot executes these trajectories and reaches
the goal without any collisions.
A group of autonomous robotic agents has more capabilities than a single robot in applications such as surveillance, information sensing, navigation, and search-and-rescue. If one can generate collision-free, non-conflict sparse paths for multiple agents and tasks at run-time, the robot swarm can explore a complex environment and execute complicated missions efficiently.

\begin{figure}[h]
	\centering
	\includegraphics[width=0.30\textwidth]{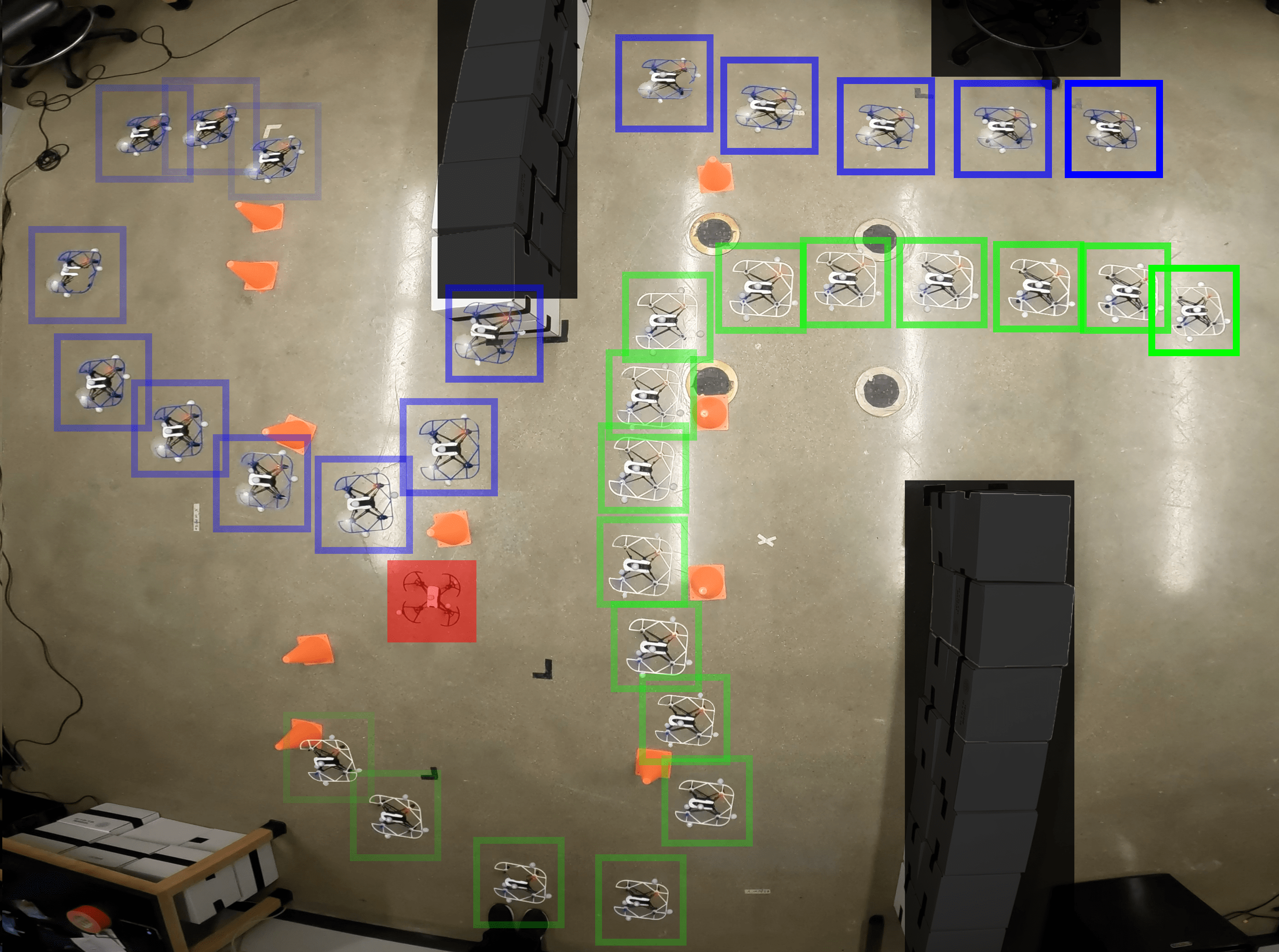}
	\caption{A screenshot of an experiment with a dynamic obstacle. Orange cones represent tasks and black areas indicate no-fly zones. A manually controlled quadrotor with red shallow represents a dynamic obstacle with infinite height. Two quadrotors with blue/green boxes are the agents and transparency indicates time.}
	\label{experiment:screenshot}
\end{figure}
In this paper, a task is defined as a location of interest that one agent must visit. Given a set of agents and tasks, a collision-aware multi-agent mission planning (MAMP) problem is defined twofold, i.e. finding optimal and conflict-free task allocations for agents and then generating collision-free paths such that agents can visit these task positions.
The former is categorized as a multi-agent task allocation (MATA) problem, and the latter is defined as a multi-agent path-finding (MAPF) problem. The optimal objective of MAPF is typically to minimize the total traveling distance. For a multi-agent system (MAS), real-time mission planning in a cluttered environment is necessary when deploying autonomous robots in a complex environment, especially when obstacles and tasks are dynamic. An example of MAMP problems is shown in Fig. \ref{example:before}.
This paper only considers MAMP problems defined as ST-SR-TA (Single-Task Robots, Single-Robot Tasks, Time-Extended Assignment) problems \cite{gerkey2004formal}. Here, tasks are assumed to be homogeneous and independent of each other, i.e. no temporal logic requirements; agents are assumed to be homogeneous regarding mission functionality. Since this problem is proven to be NP-hard \cite{gerkey2004formal}, there is a trade-off for MAMP problems between real-time performance and optimality. Furthermore, the scalability of an underlying algorithm, in terms of the number of agents and tasks, is crucial in MAS applications.



\subsection{Related Work}\label{subsec:related_work}

The literature on MAMP problems basically can be divided into two categories, i.e. solving MATA and MAPF problems sequentially or in an integrated way.

The methods related to MATA can be mainly categorized as auction-based and searching-based methods.
Auction-based approaches are derived from a concept in finance where each agent aims to maximize their own reward by giving higher bids. And the process must consider maximizing a global reward and include conflict resolution.
\cite{michael2008distributed} utilizes auction-based protocols to bid task assignments.
CBBA (Consensus-Based Bundle Algorithm)\cite{choi2009consensus} employs a decentralized consensus procedure for task conflict resolution and then generates task allocation for agents.
IACA (Iterated Auction Consensus Algorithm) \cite{wang2021distributed} proposes a similar iterative but resilient auction process and can remove malicious bids during the auction.
\cite{nunes2017decentralized} proposed an auction-based algorithm to deal with task allocation problems with time window constraints.
\cite{kim2019minimizing} produces task sequences with minimum communications by combining the greedy algorithm and the auction process.
Although the auction-based approaches are decentralized, the process of auction and conflict resolution can be time-consuming, especially when the problem size is large. In addition, the auction heuristic barely includes environmental information, e.g. the impact of obstacles on the cost/reward. Thus, the auction result is not necessarily optimal when the obstacles are present and may even lead to a bad solution.

Search-based methods rely on a fixed structure of information, e.g. the number of assigned tasks for each agent is known and fixed.
\cite{patel2020decentralized} proposes a decentralized genetic algorithm (GA) to search a task sequence parallelly. 
\cite{banks2020multi} proposes a graph-based search method to allocate tasks to agents given a finite linear temporal logic objective, where the allocation order is partially known.
\cite{henkel2020optimized} builds an Optimized Directed Roadmap Graph (ODRM) by sampling first, and then navigates agents on this graph. Although searching paths on an ODRM is faster than on the most common occupancy grid map, generating and updating such a graph at run-time can be time-consuming in a cluttered and dynamic environment.


Due to the page limit, this paper omits the literature on MAPF problems because most of the recent literature focuses on the integration of MATA and MAPF problems. As for the literature on solving MATA and MAPF problems sequentially, they are mainly categorized as auction-based and search-based methods. 
Based on CBBA, \cite{bertuccelli2009real} first generates task sequences without any obstacle information and then utilizes Dijkstra's algorithm \cite{dijkstra1959note} to find collision-free paths given the sequences.
\cite {choi2011genetic} proposes a two-stage GA-based approach where each agent first determines its own task sequence using a genetic algorithm and then negotiates with other agents to exchange tasks if that reduces the cost. Then collision-free paths are generated similarly as \cite{bertuccelli2009real}.

There are also some special cases of MAMP problems that have risen significant interest, such as multi-agent pickup and delivery\cite{henkel2019optimal}, and vehicle routing problems. Some special specifications are adopted for these problems. For example, the task set for each agent is prescribed; each agent can only be assigned one task; the initial positions for agents are the same, etc. This paper considers a general MAMP problem without these special specifications.

There is some literature on the integrated MAMP methods.
\cite{schillinger2018simultaneous} focuses on simultaneous task allocation and planning for a complex goal that consists of temporal logic sub-tasks. \cite{schillinger2018simultaneous} emphasizes the capability of a heterogeneous robot team to perform a complex goal, whereas the MAMP problem in this paper focuses on homogeneous agents and tasks.
\cite{ren2021ms}, as a fully centralized optimization-based method, first obtains a single tour that connects all the tasks without any obstacle information by solving a traveling salesman problem; then uses a heuristic policy to partition the tour to generate a task allocation sequence for each agent; finally generates collision-free paths. Although \cite{ren2021ms} deals with the same problem with this paper, its computation time is stably around 55 seconds, with 5 - 20 agents and 10 - 50 tasks in a map with random obstacles.

From the methodology perspective, there are primarily three types of methods for MAMP problems with homogeneous agents/tasks and no temporal logic constraints, i.e. decentralized auction-based, distributed GA-based (genetic algorithm), and centralized optimization-based methods. Decentralized auction-based methods, as mentioned above, suffer from inefficient auction and negotiation processes and a lack of obstacle information during the auction process. Distributed GA-based methods might have good real-time performance for small-size problems but it notably depends on the selection of GA parameters. Also, many methods assume the number of assigned tasks for each agent is known and fixed, whereas this paper does not.
As for optimization-based methods, they barely utilize obstacle information in the first place and are not in a distributed manner, i.e directly solving the entire allocation problem.

\subsection{Contributions and Notations}\label{subsec:contribution}
This paper proposes a real-time MAMP algorithm DrMaMP for homogeneous agents and tasks. DrMaMP first utilizes obstacle information as heuristics to approximate the cost of an ordered task allocation and path sequence by a metric from an unordered set. With this approximation, DrMaMP can partition the entire problem into several sub-problems and distribute them to each agent. Then each agent finds optimal task allocation and path sequence for each sub-problem. Due to the approximation and the distributed manner, DrMaMP makes a balance between computational performance and scalability.
The main contributions are:
\begin{enumerate}
	\item a distributed real-time (on the order of millisecond) MAMP algorithm DrMaMP;
	\item capability of handling dynamic obstacles and tasks in a cluttered environment at run-time;
	\item good scalability in terms of the number of agents and tasks and relatively good optimality;
	\item computational burden analysis for DrMaMP.
\end{enumerate}


\textit{Notations.} Vectors, variables, and functions in multiple dimensions are in bold lowercase; matrices and sets are in uppercase. For a point $\boldsymbol{p} \in \mathbb{R}$, $\{ \boldsymbol{p} \} \subset \mathbb{R}$ denotes a set containing that point as its only element. Set subtraction is ${A \setminus B=\{x\in A\mid x\notin B\}}$.
$\mathbb{Z}$ denotes the integer set.
$\mathbb{Z}_{+}$ denotes the positive integer set.
The cardinality of a set $A$ is denoted as $|A|$.


\section{Problem Formulation}\label{sec:problem_formulation}

The configuration space, $X \subseteq \mathbb{R}^n$, is all positions in space reachable by an agent. Denote an agent positions set $\mathcal{X} = \{\boldsymbol{p}_1,\cdots,\boldsymbol{p}_{n_a}\}$ of $n_a$ agents, and $\boldsymbol{p}_i \in X$ is the position of agent $i$. Denote a task positions set $\mathcal{T} = \{\boldsymbol{t}_1,\cdots,\boldsymbol{t}_{n_t}\}$ of $n_t$ tasks, and $\boldsymbol{t}_i \in X$ is the position of task $i$.
Define the agent and tasks index sets $\mathcal{I} \triangleq \{1,\cdots,n_a\}$ and $\mathcal{J} \triangleq \{1,\cdots,n_t\}$, respectively.
Suppose that an agent completes a task when the distance between two entities is less than a prescribed non-negative constant $\epsilon$, i.e. $||\boldsymbol{p}_i - \boldsymbol{t}_j||_2 \leq \epsilon, \ \epsilon \geq 0$.
Denote an obstacle positions set as $\mathcal{O} = \{ \boldsymbol{o}_1, \cdots, \boldsymbol{o}_{n_{o}} \}$ of $n_o$ obstacles, where $\boldsymbol{o}_i \in X$ is the position of obstacle $i$. Denote $\mathcal{P}_i \triangleq (\boldsymbol{p}^0_i, \boldsymbol{p}^1_i, \cdots, \boldsymbol{p}^{n_{p,i}-1}_i) \subset X$ as an ordered sequence of positions associated with agent $i$ which denotes a path starting from $\boldsymbol{p}^0_i$ and ending at $\boldsymbol{p}^{n_{p,i}-1}_i$, where $n_{p,i} \triangleq |\mathcal{P}_i|$ denotes the number of positions in $\mathcal{P}_i$.

Derived from \cite{choi2009consensus}, the collision-aware MATA problem is written as the following integer programming:
\begin{mini!}|s|[2]
	{ \boldsymbol{x}, \boldsymbol{r}_1, \cdots, \boldsymbol{r}_{n_a} }{\textstyle \sum_{i=1}^{n_a} \sum_{j=1}^{n_t} c_{ij}(\boldsymbol{x}_i, \boldsymbol{r}_i, \mathcal{O})x_{ij} \label{task_allocation_prob:obj}}
	{\label{task_allocation_prob}}{}
	\addConstraint{\textstyle \sum_{j=1}^{n_t}x_{ij} \leq n_t, \ \forall i \in \mathcal{I}}{ \label{task_allocation_prob:task_constraint}}
	\addConstraint{\textstyle \sum_{i=1}^{n_a}x_{ij} = 1, \ \forall j \in \mathcal{J}}{ \label{task_allocation_prob:agent_constraint}}
	\addConstraint{\textstyle \sum_{i=1}^{n_a} \sum_{j=1}^{n_t} x_{ij} = n_t }{ \label{task_allocation_prob:complete_constraint}}
	\addConstraint{ x_{ij} \in \{0,1\}, \ \forall (i,j) \in \mathcal{I} \times \mathcal{J},}{ \label{task_allocation_prob:decision_variable}}
\end{mini!}
where $x_{ij}=1$ if task $j$ is assigned to agent $i$ and $0$ otherwise; $\boldsymbol{x}_i \in \{0,1\}^{n_t}$ is the task assignment vector for agent $i$, $x_{ij}$ is the $j$-th element of $\boldsymbol{x}_i$, and $\boldsymbol{x} = \matt{ \boldsymbol{x}_1^{\prime} & \cdots & \boldsymbol{x}_{n_a}^{\prime} }^{\prime} \in \{0,1\}^{n_t n_a}$.
The vector $\boldsymbol{r}_i \in \{\mathcal{J} \cup \{\emptyset\}\}^{n_a}$ denotes an ordered sequence of tasks, i.e., the task allocation order, for agent $i$; its $k$-th element is $j \in \mathcal{J}$ if task $j$ is the $k$-th task of agent $i$'s assignment; $\boldsymbol{r}_i = \emptyset$ if agent $i$ has no assignment.
The collision-aware cost of task $j$ being assigned to agent $i$ followed by an order $\boldsymbol{r_i}$ is defined by $c_{ij}(\boldsymbol{x}_i, \boldsymbol{r}_i, \mathcal{O}) \geq 0$. In the context of mission planning, this cost typically represents traveling distance, fuel consumption, etc.
Constraint \eqref{task_allocation_prob:task_constraint} indicates that each agent can be at most assigned with $n_t$ tasks;
\eqref{task_allocation_prob:agent_constraint} requires that each task must be assigned to only one agent;
\eqref{task_allocation_prob:complete_constraint} enforces that every task must be assigned.

Denote a task allocation order set $\mathcal{R} \triangleq \{ \boldsymbol{r}_1, \cdots, \boldsymbol{r}_{n_a} \}$.
Given an order set $\mathcal{R}$ and the current positions of agents $\mathcal{X}$, the collision-aware MAPF problem is written as:
\begin{mini}|s|
	{\mathcal{P}_1,\cdots,\mathcal{P}_{n_a}}{\textstyle \sum_{i=1}^{n_a} \ell_i(\mathcal{P}_i)}
	{\label{multi_agent_path_planning}}{}
	\addConstraint{\boldsymbol{p}_i^0 = \boldsymbol{p}_i, \ \forall i \in \mathcal{I}}
	\addConstraint{\mathcal{R} \text{ is determined by } \eqref{task_allocation_prob}}
	\addConstraint{\mathcal{P}_i \text{ satisfies the order } \boldsymbol{r}_i, \ \forall i \in \mathcal{I}}
	\addConstraint{\mathcal{P}_i \cap \mathcal{O} = \emptyset, \ \forall i \in \mathcal{I},}
\end{mini}
where $\ell_i(\mathcal{P}_i) = \sum_{j=0}^{|\mathcal{P}_i|-2} ||\boldsymbol{p}_i^{j+1} - \boldsymbol{p}_i^{j}||_2$ is the traveling distance of path $\mathcal{P}_i$. This paper assumes that $\mathcal{P}_i \cap \mathcal{O} = \emptyset$ if and only if $||\boldsymbol{p}_i^j - \boldsymbol{o}_k||_2 \geq \delta > 0 \ \forall \boldsymbol{p}_i^j \in \mathcal{P}_i \text{ and } \forall \boldsymbol{o}_k \in \mathcal{O}$.

Based on \eqref{task_allocation_prob} and \eqref{multi_agent_path_planning}, the collision-aware MAMP problem in this paper is formulated as:
\begin{mini}|s|
	{\boldsymbol{x}, \mathcal{R}, \mathcal{P}}{\textstyle \sum_{i=1}^{n_a} \sum_{j=1}^{n_t} c_{ij}(\boldsymbol{x}_i, \boldsymbol{r}_i, \mathcal{O})x_{ij}}
	{\label{problem_this}}{}
	\addConstraint{\textstyle \sum_{j=1}^{n_t} x_{ij} \leq n_t, \ \forall i \in \mathcal{I}}
	\addConstraint{\textstyle \sum_{i=1}^{n_a} x_{ij} = 1, \ \forall j \in \mathcal{J}}
	\addConstraint{\textstyle \sum_{i=1}^{n_a} \sum_{j=1}^{n_t} x_{ij} = n_t}
	\addConstraint{ x_{ij} \in \{0,1\}, \ \forall (i,j) \in \mathcal{I} \times \mathcal{J}}
	\addConstraint{\mathcal{P} \triangleq \{ \mathcal{P}_1, \cdots, \mathcal{P}_{n_a} \} \text{ is determined by } \eqref{multi_agent_path_planning}}
	\addConstraint{\mathcal{R} \text{ is determined by } \eqref{task_allocation_prob},}
\end{mini}
where $\textstyle \sum_{j=1}^{n_t} c_{ij}(\boldsymbol{x}_i, \boldsymbol{r}_i, \mathcal{O})x_{ij}$ evaluates agent $i$'s collision-aware traveling distance given a particular assignment and allocation order.

Solving the task assignment $\boldsymbol{x}$, the allocation order $\mathcal{R}$, and the collision-free path $\mathcal{P}$ altogether is challenging because $\boldsymbol{x}$, $\mathcal{R}$, and $\mathcal{P}$ are coupled together in \eqref{task_allocation_prob}, \eqref{multi_agent_path_planning}, and \eqref{problem_this}.
Furthermore, the collision-aware MAMP problem \eqref{problem_this} is not even tractable since it is proven to be NP-hard \cite{gerkey2004formal}. This paper attempts to obtain a sub-optimal solution to the collision-aware MAMP problem scalably and in real-time, especially when the environment is unconstructed and cluttered and the obstacles and tasks are potentially dynamic.

\section{Algorithm}\label{sec:approach}
This paper proposes a Distributed Real-time Multi-agent Mission Planning (DrMaMP) algorithm to obtain a sub-optimal solution to \eqref{problem_this} in a scalable way.
Instead of considering the exact coupled cost $c_{ij}(\boldsymbol{x}_i, \boldsymbol{r}_i, \mathcal{O})$, DrMaMP utilizes task-based heuristics to approximate the cost of an ordered path by an unordered set. With this approximation, DrMaMP can partition the entire task set into several subsets and assign each task subset to one agent given the unordered heuristics. Then each agent only needs to solve a sub-problem, i.e. single-agent mission planning problem.
Specifically, DrMaMP consists of three phases:
\begin{enumerate}[noitemsep,nolistsep]
	\item Task Segmentation: partitioning the entire task set into several subsets;
	\item Cluster Assignment: assigning each agent a task subset;
	\item Single-Agent Mission Planning: finding an optimal task allocation order and collision-free path for each agent.
\end{enumerate}

In Phase 1, given an objective defined in Section \ref{subsec:task_segmentation}, the entire task set $\mathcal{T}$ is partitioned into $n_a$ subsets.
In Phase 2, given an objective defined in Section \ref{subsec:cluster_assign}, each agent is assigned one task subset by solving an assignment problem.
In Phase 3, after each agent is assigned with a task subset, it needs to solve a single-agent mission planning problem individually to find the optimal task allocation order and collision-free path. The computation can be distributed to each agent.
A detailed explanation of the 3 phases is shown in the following subsections.

\begin{figure*}[h]
	\centering
	\begin{subfigure}{.25\textwidth}
		\centering
		\includegraphics[width=\linewidth]{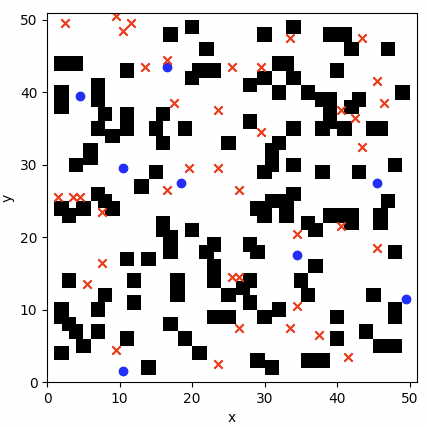}
		\caption{An example problem}
		\label{example:before}
	\end{subfigure}
	\hfill
	\begin{subfigure}{.25\textwidth}
		\centering
		\includegraphics[width=\linewidth]{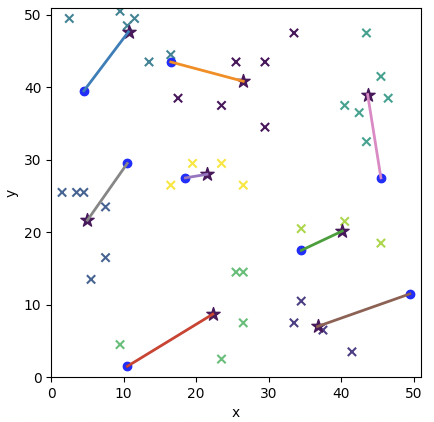}
		\caption{Task segmentation and cluster assignment result}
		\label{example:cluster}
	\end{subfigure}
	\hfill
	\begin{subfigure}{.25\textwidth}
		\centering
		\includegraphics[width=\linewidth]{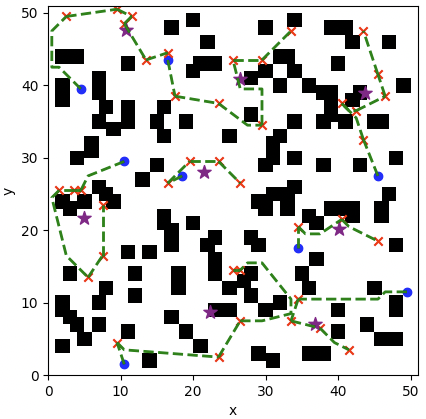}
		\caption{Mission planning result}
		\label{example:after}
	\end{subfigure}
	\caption{ An example for MAMP with 8 agents and 40 tasks. (a) shows the problem to be solved in a $50 \times 50$ grid map with 150 obstacles, where the blue dots and red crosses indicate the positions of agents and tasks, respectively; (b) shows the task segmentation and cluster assignment result, where those tasks in the same color are within the same cluster, the purple stars indicate the positions of cluster centroids and an edge between an agent and a cluster centroid represents assignment; (c) shows the task allocation orders and collision-free paths, where the dashed lines in green indicate the paths. $N=300$ in Algorithm \ref{alg:algo_task_segment}. The computation time is 44.6 ms.} 
	\label{comp:fixed_target}
\end{figure*}


\subsection{Task Segmentation}\label{subsec:task_segmentation}
The entire task set $\mathcal{T}$ is partitioned into $n_a$ clusters $\{\mathcal{T}_1, \cdots, \mathcal{T}_{n_a}\}$, where each cluster includes possibly many tasks. Note that $\mathcal{T}_i$ has not been assigned to any agents yet. The tasks within a cluster have a minimal distance to the centroid of this cluster. Such a task segmentation can be obtained by an iterative k-means clustering algorithm \cite{lloyd1982least}, which minimizes the summation of the within-cluster sum of squares (WCSS), i.e.
\begin{equation}
	\label{eq:problem_k_means}
	\begin{aligned}
		\min_{\mathcal{T}_1, \cdots, \mathcal{T}_{n_a}} \quad & \textstyle \sum_{i=1}^{n_a} \sum_{\boldsymbol{t} \in \mathcal{T}_i} ||\boldsymbol{t}-\boldsymbol{c}_i||_2^2 \\
		\textrm{s.t.} \quad
		& \mathcal{T} = \cup_{i=1}^{n_a} \mathcal{T}_i, \\
		& \mathcal{T}_i \cap \mathcal{T}_j = \emptyset, \ \forall i \neq j, \\
		& \boldsymbol{c}_i =  (\Sigma_{\boldsymbol{t} \in \mathcal{T}_i} \  \boldsymbol{t}) / |\mathcal{T}_i|, \ \forall i,
	\end{aligned}
\end{equation}
where $\mathcal{T}_i = \{ \boldsymbol{t}_j \ | \ \forall j \in \mathcal{I}_{c,i} \}$ and $\mathcal{I}_{c,i}$ is the task index set that is associated with the tasks within cluster $\mathcal{T}_i$; $\boldsymbol{c}_i \in \mathbb{R}^n$ is the centroid of tasks within $\mathcal{T}_i$. Denote $\mathcal{C} \triangleq \{ \boldsymbol{c}_1, \cdots, \boldsymbol{c}_k \}$.

As described in \eqref{problem_this}, the objective is to minimize the total traveling distance. But the cost of each agent visiting a known task set is unknown before a task allocation order is determined. Hence, for each task subset $\mathcal{T}_i$, an ordered sequence's length is approximated by an unordered set's WCSS, i.e. $\sum_{\boldsymbol{t} \in \mathcal{T}_i} ||\boldsymbol{t}-\boldsymbol{c}_i||_2^2$, since the tasks within $\mathcal{T}_i$ have a less WCSS associated with $\boldsymbol{c}_i$ than $\boldsymbol{c}_j \ \forall j \neq i$. The task segmentation problem \eqref{eq:problem_k_means} can be solved iteratively and the details are in Algorithm~\ref{alg:algo_task_segment}. An example is shown in Fig. \ref{example:cluster}.

\subsection{Cluster Assignment}\label{subsec:cluster_assign}
Since an ordered sequence's length is approximated by an unordered set's WCSS in Algorithm \ref{alg:algo_task_segment}, the cost of agent $i$ visiting a task subset is approximated by the distance between the agent and task subset's centroid plus the subset's WCSS, which is independent on the task allocation order.
Then the task subset (cluster) assignment problem is written as an integer linear programming \eqref{eq:problem_cluster_assignment},
where $y_{ij}=1$ if agent $i$ is assigned with cluster $j$ and $0$ otherwise; $w_{ij} \triangleq ||\boldsymbol{p}_i-\boldsymbol{c}_j||^2_2 + \sum_{\boldsymbol{t} \in \mathcal{T}_j} || \boldsymbol{t}-\boldsymbol{c}_j ||^2_2$ defines the cost of cluster $j$ being assigned to agent $i$, where the first term evaluates how far agent $i$ is to cluster $j$ and the second term estimates the cost of agent $i$ visiting all the tasks within cluster $j$.
\begin{mini!}|s|[2]
	{\boldsymbol{y}}{\textstyle \sum_{i \in \mathcal{I}} \sum_{j \in \mathcal{I}_{c,j}} w_{ij}y_{ij} \label{eq:problem_cluster_assignment:obj}}
	{\label{eq:problem_cluster_assignment}}{}
	\addConstraint{\textstyle \sum_{i \in \mathcal{I}} y_{ij} = 1, \ \forall j \in \mathcal{I}_{c,j}}{\label{eq:problem_cluster_assignment:agent_constraint}}
	\addConstraint{\textstyle \sum_{j \in \mathcal{I}_{c,j}} y_{ij} \leq 1, \ \forall i \in \mathcal{I}}{\label{eq:problem_cluster_assignment:cluster_constraint}}
	\addConstraint{\textstyle \sum_{i \in \mathcal{I}}\sum_{j \in \mathcal{I}_{c,j}} y_{ij} = n_a}{\label{eq:problem_cluster_assignment:complete_constraint}}
	\addConstraint{y_{ij}=\{0,1\}, \ \forall (i,j) \in \mathcal{I} \times \mathcal{I}_{c,j}.}{\label{eq:problem_cluster_assignment:decision_variable_constraint}}
\end{mini!}
Constraint \eqref{eq:problem_cluster_assignment:agent_constraint} ensures that each cluster must be assigned with one agent; \eqref{eq:problem_cluster_assignment:cluster_constraint} guarantees that each agent can be at most assigned to one cluster; \eqref{eq:problem_cluster_assignment:complete_constraint} enforces no unassigned cluster left.
Constraint \eqref{eq:problem_cluster_assignment:cluster_constraint} considers a situation when the number of agents is greater than the number of nonempty clusters. This situation can happen at run-time when some tasks are completed.
The cluster assignment problem \eqref{eq:problem_cluster_assignment} can be solved by some constrained integer linear programming solvers such as OR-Tools \cite{ortools}.

\begin{algorithm}
	\caption{Task Segmentation}\label{alg:algo_task_segment}
	\DontPrintSemicolon
	\KwIn{$\mathcal{T}$, $N \in \mathbb{Z}_{+}$}
	\KwOut{$\{\mathcal{T}_1, \cdots, \mathcal{T}_{n_a}\}$, $\{ \mathcal{I}_{c,1}, \cdots, \mathcal{I}_{c,n_a} \}$, $\mathcal{C}$}
	
	Initialize \textbf{Output} by k-means++ \cite{arthur2006k},  $iter = 0$\;
	\While {$iter < N$} {
		\For {$\text{task } \boldsymbol{t}_i = \boldsymbol{t}_1 \ \text{to} \  \boldsymbol{t}_{n_t}$} {
			$idx$ $\gets$ the index of $t_i$'s nearest centroid \;
			$\mathcal{I}_{c,idx}.$append($i$)\;
		}
		
		\For {$j = 1 \  \text{to} \  n_a$} {
			$\boldsymbol{c}_j$ $\gets$ mean of all tasks within cluster $j$\;
		}
		
		$iter \gets iter + 1$\;
	}
	
	\lFor {$i = 1 \  \text{to} \  n_a$} {
		$\mathcal{T}_i \gets \{\boldsymbol{t}_j \ | \  \forall j \in \mathcal{I}_{c, i} \}$
	}
	
	\Return $\{\mathcal{T}_1, \cdots, \mathcal{T}_{n_a}\}$, $\{ \mathcal{I}_{c,1}, \cdots, \mathcal{I}_{c,n_a} \}$, $\mathcal{C}$
\end{algorithm}

\begin{algorithm}
	\caption{Cluster Assignment}\label{alg:cluster_assignment}
	\DontPrintSemicolon
	\KwIn{$\{\mathcal{T}_1, \cdots, \mathcal{T}_{n_a}\}$, $\mathcal{X}$, $\mathcal{C}$}
	\KwOut{$\{\hat{\mathcal{T}}_1, \cdots, \hat{\mathcal{T}}_{n_a}\}$}
	
	\For {$\text{agent } i = 1 \ \text{to} \  n_a$} {
		\For {$\text{cluster } j = 1 \ \text{to} \  n_a$} {
			$w_{ij} \gets ||\boldsymbol{p}_i-\boldsymbol{c}_j||^2_2 + \sum_{\boldsymbol{t} \in \mathcal{T}_j} || \boldsymbol{t}-\boldsymbol{c}_j ||^2_2$\;
		}
	}
	$\boldsymbol{y}^* \gets$ Solve \eqref{eq:problem_cluster_assignment} by an numerical solver\;
	$\{\hat{\mathcal{T}}_1, \cdots, \hat{\mathcal{T}}_{n_a}\} \gets$ parse\_result($\{\mathcal{T}_1, \cdots, \mathcal{T}_k\}, \  \boldsymbol{y}^*$)\;
	
	\Return $\{\hat{\mathcal{T}}_1, \cdots, \hat{\mathcal{T}}_{n_a}\}$
\end{algorithm}

Denote that $\hat{\mathcal{T}}_i$ is the task cluster assigned to agent $i$. Details about the cluster assignment are shown in Algorithm~\ref{alg:cluster_assignment}. An example is shown in Fig. \ref{example:cluster}.

\subsection{Single-Agent Mission Planning}

After each agent is assigned a task cluster, the task allocation orders and the collision-free paths need to be determined. This problem can be distributed to $n_a$ agents parallelly, and agent $i$ solves its own sub-problem by formulating it as a Travelling Salesperson Problem (TSP), where the nodes are the agent itself and its assigned tasks. A path-finding algorithm generates collision-free paths for every pair of nodes and the length of these paths is the traveling cost from one node to another. This paper utilizes Lazy Theta* \cite{nash2010lazy} as the path-finding algorithm due to fewer line-of-sight checks.
The single-agent mission planning problem can be written as an integer linear program \eqref{eq:tsp:obj} with Miller–Tucker–Zemlin (MTZ) formulation \cite{miller1960integer},
\begin{mini!}|s|[2]
	{\boldsymbol{z},\boldsymbol{u}}{\textstyle \sum_{i=1}^{n_m} \sum_{j=1}^{n_m} d_{ij}z_{ij} \label{eq:tsp:obj}}
	{\label{eq:tsp}}{}
	\addConstraint{z_{ij} \in \{0,1\}, \ \forall i,j=1,\cdots,n_m}{\label{eq:tsp:decision_variable}}
	\addConstraint{u_i \in \mathbb{Z}, \ \forall i=2,\cdots,n_m}{\label{eq:tsp:dummpy_variable}}
	\addConstraint{u_i-u_j+n_m z_{ij} \leq n_m-1, \ 2 \leq i \neq j \leq n_m}{\label{eq:tsp:sub_tour_constraint}}
	\addConstraint{1 \leq u_i \leq n_m-1}{\label{eq:tsp:sub_tour_bound}}
	\addConstraint{\textstyle \sum_{i=1}^{n_m} z_{ij}=1, \ \forall j=2,\cdots,n_m}{\label{eq:tsp:start_constraint}}
	\addConstraint{\textstyle \sum_{j=1}^{n_m} z_{ij}=1, \ \forall i=1,\cdots,n_m}{\label{eq:tsp:end_constraint}}
	\addConstraint{\textstyle \sum_{i=1}^{n_m} z_{i1}=0,}{\label{eq:tsp:no_go_back_home_constraint}}
\end{mini!}
where $n_m \triangleq |\hat{\mathcal{T}}_i| + 1$ denotes the number of nodes; node 1 always indicates the agent's current position; $z_{ij}=1$ if the agent goes from node $i$ to node $j$, $\boldsymbol{z} \in \{0,1\}^{n_m^2}$; $\boldsymbol{u} \in \mathbb{Z}^{n_m-1}$ is a dummy variable to indicate tour ordering such that $u_i < u_j$ implies node $i$ is visited before node $j$; $d_{ij}$ is the cost of agent traveling from node $i$ to node $j$, which is the length of the underlying collision-free path.
\begin{algorithm}
	\caption{Task Allocation and Path Finding}\label{alg:single}
	\DontPrintSemicolon
	\KwIn{$\{\hat{\mathcal{T}}_1, \cdots, \hat{\mathcal{T}}_{n_a}\},\mathcal{X},\mathcal{O}$}
	\KwOut{$\{\mathcal{P}_1,\cdots,\mathcal{P}_{n_a}\}, \{\boldsymbol{r}_1,\cdots,\boldsymbol{r}_{n_a}\}$}
	
	// $n_a$ agents parallelly execute the content in \textbf{parfor}\;
	$\textbf{par}$\For {agent $i = 1 \ to \ n_a$} {
		Initialize $P_{lib}$ as empty\;
		\For {start, goal in $(\hat{\mathcal{T}}_i \cup \{\boldsymbol{p}_i\})$} {
			$\mathcal{O}_{now} \gets \mathcal{O} \cup \mathcal{X} \setminus \{\boldsymbol{p}_i\}$\;
			$P_{start,goal} \gets$ path\_finding($start, goal, \mathcal{O}_{now}$)\;
			$P_{lib}.$append($P_{start,goal}$)\;
		}
		
		\For {$\text{node } i = 1 \ \text{to} \  1+|\hat{\mathcal{T}}_i|$} {
			\For {$\text{node } j = 1 \ \text{to} \  1+|\hat{\mathcal{T}}_i|$} {
				$P_{i,j} \gets $load\_path($P_{lib},i,j$)\;
				$d_{ij} \gets $compute\_cost($P_{i,j}$)\;
			}
		}
		$\boldsymbol{z}^*, \boldsymbol{u}^* \gets$ solve \eqref{eq:tsp} by a numerical solver\;
		$\mathcal{P}_i, \boldsymbol{r}_i \gets$ parse\_path($P_{lib}, \boldsymbol{z}^*, \boldsymbol{u}^*$)\;
	}
	\Return $\{\mathcal{P}_1,\cdots,\mathcal{P}_{n_a}\}, \{\boldsymbol{r}_1,\cdots,\boldsymbol{r}_{n_a}\}$
\end{algorithm}
The constraints \eqref{eq:tsp:dummpy_variable} - \eqref{eq:tsp:sub_tour_bound} guarantees only one tour covering all nodes \cite{miller1960integer}. The constraints \eqref{eq:tsp:start_constraint} - \eqref{eq:tsp:end_constraint} guarantees that each node is visited from another node, and from each node, there is a departure to another node. The constraint \eqref{eq:tsp:no_go_back_home_constraint} indicates that the agent does not go back to its initial position after visiting all the tasks. One can change \eqref{eq:tsp:no_go_back_home_constraint} if the agent needs to go back to a base. To ensure that there is no collision between agents, each agent considers the other agents as obstacles.
Details are shown in Algorithm \ref{alg:single}. An example is shown in Fig. \ref{example:after}.

\subsection{DrMaMP at Run-time}\label{subsec:DrMaMP_in_realtime}
This subsection illustrates how DrMaMP operates at run-time. First, DrMaMP utilizes k-means++ \cite{arthur2006k} to initialize the cluster centroids. During the mission, the centroids from the previous timestamp are the initial centroids for the next timestamp. As some tasks are completed, the number of nonempty clusters $n_c$ might be less than $n_a$. If $n_c < n_a$, one needs to remove the empty clusters and revise the constraint \eqref{eq:problem_cluster_assignment:complete_constraint} as $\sum_{i \in \mathcal{I}} \sum_{j \in \mathcal{I}_{c,j}} y_{ij}=n_c$.
Note that all the constraints are compatible with the case where $n_c < n_a$.
If there exist dynamic obstacles and tasks, DrMaMP updates their information (positions) in each timestamp. More details are shown in Algorithm \ref{alg:DrMaMP}.

\begin{algorithm}
	\caption{DrMaMP at Run-time}\label{alg:DrMaMP}
	\DontPrintSemicolon
	
	Initialize $\mathcal{C}$ by k-means++ \cite{arthur2006k}\;
	\While {$\mathcal{T} \neq \emptyset$} {
		$\mathcal{T} \gets$ update task set \;
		$\mathcal{X} \gets$ update agent position \;
		$\mathcal{O} \gets$ update obstacle \;
		
		$\{\mathcal{T}_1, \cdots, \mathcal{T}_k\}, \mathcal{C} \gets$ Algorithm \ref{alg:algo_task_segment} with previous $\mathcal{C}$ \;
		
		remove empty task cluster \;
		
		$\{\hat{\mathcal{T}}_1, \cdots, \hat{\mathcal{T}}_{n_a}\} \gets$ Algorithm \ref{alg:cluster_assignment} \;
		
		$\{\mathcal{P}_1,\cdots,\mathcal{P}_{n_a}\}, \{\boldsymbol{r}_1,\cdots,\boldsymbol{r}_{n_a}\} \gets$ Algorithm \ref{alg:single} \;
		
		agents move one step along $\{\mathcal{P}_1,\cdots,\mathcal{P}_{n_a}\}$ \;
		time moves one step forward \;
		
		$\boldsymbol{t}_j \gets$ current assigned task of agent $i, \  \forall i \in \mathcal{I}$ \;
		delete task $\boldsymbol{t}_j$ if $||\boldsymbol{p}_i-\boldsymbol{t}_j||_2 \leq \epsilon, \  \forall i \in \mathcal{I}$ \;
		
	}

\end{algorithm}



\section{Comparisons and Experiments}\label{sec:results}
This section presents several experiments with static/dynamic obstacles/tasks and conducts scalability and optimality comparisons between DrMaMP and a decentralized method \cite{bertuccelli2009real}. From here on, CBBA is interchangeable with the method in \cite{bertuccelli2009real} because it consists of CBBA and posterior path-finding. DrMaMP outperforms \cite{bertuccelli2009real} in both indices based on the comparisons. In addition, this section analyzes the computational burden for DrMaMP and presents the optimality gap in small-size problems.

DrMaMP is written in C++ and compiled as a Python library to be invoked. The integer programmings in Algorithm \ref{alg:cluster_assignment} and \ref{alg:single} are solved by Google OR-Tools\cite{ortools}. The C++ implementation utilizes multithreading to simulate the distributed manner, i.e., the \emph{parfor} in Line 2, Algorithm \ref{alg:single}. First, a main thread, i.e. the central agent, runs Algorithm \ref{alg:algo_task_segment} and Algorithm \ref{alg:cluster_assignment}. Then the results of Algorithm \ref{alg:cluster_assignment} are distributed to multiple agents/threads, where each thread runs Algorithm \ref{alg:single} parallelly for each agent.
All the results are obtained by a desktop with a 2.8 GHz Intel Core i7-7700HQ CPU and 16 GB memory.
This implementation does not require a GPU but one can accelerate it with GPU if needed.


\subsection{Experiments} \label{subsec:experiments}
The test area is 6m $\times$ 5.6m and the grid map size is $120 \times 112$. DrMaMP runs on the same desktop and does real-time mission planning for two Parrot Mambo quadrotors. In the experiments, each quadrotor follows the discrete paths returned from DrMaMP.
Then a low-level trajectory tracking controller\footnote[1]{ \url{github.com/zehuilu/Mambo-Tracking-Interface} } broadcasts desired control commands given the desired paths to each Mambo individually.
Some details are explained in Fig. \ref{experiment:screenshot}.
In the case of a dynamic task, a cone moves from one side to another side and DrMaMP updates its planning result accordingly.
The video also includes simulations with many dynamic obstacles in a cluttered environment.
Footage from these experiments is included in a supplementary video file\footnote[2]{\url{youtu.be/il3YxhXgGac}}.

\subsection{Comparison with Increased Number of Agents} \label{subsec:comp_agen}
Section \ref{subsec:comp_agen} and Section \ref{subsec:comp_task} show scalability comparisons between DrMaMP and \cite{bertuccelli2009real}. The grid map is $50 \times 50$. 
Given a particular number of agents $n_a$ and tasks $n_t$, there are 100 different scenarios where the positions of agents and tasks are generated randomly. For each scenario, there are 200 randomly generated obstacles; each method runs 20 times, and the average computation time and total distance are collected.

Although \cite{bertuccelli2009real} utilizes Dijkstra's algorithm \cite{dijkstra1959note} as the path-finder, this paper replaces Dijkstra's algorithm by Lazy Theta* \cite{nash2010lazy} as the path-finder of \cite{bertuccelli2009real} to present a fair comparison, regarding the computation. In other words, this paper cancels the difference between two path-finders although Lazy Theta* is faster, occupies less memory, and generates shorter paths due to any-angle movement.

\begin{figure*}[h]
	\centering
	\begin{subfigure}{.30\textwidth}
		\centering
		\includegraphics[width=\linewidth]{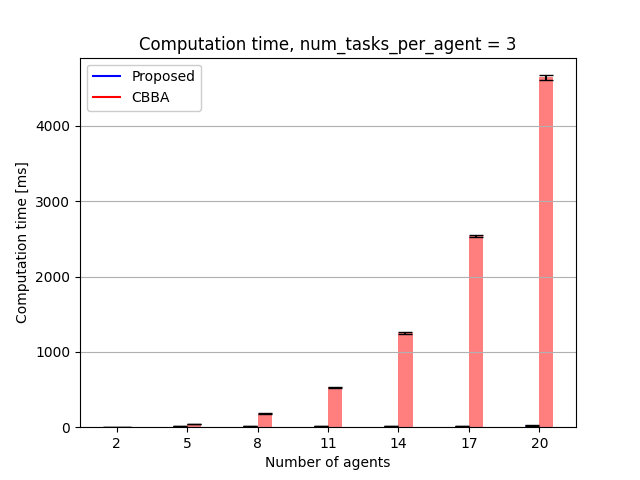}
		\caption{Computation time of two methods}
		\label{comp:fixed_target:time_two}
	\end{subfigure}
	\hfill
	\begin{subfigure}{.30\textwidth}
		\centering
		\includegraphics[width=\linewidth]{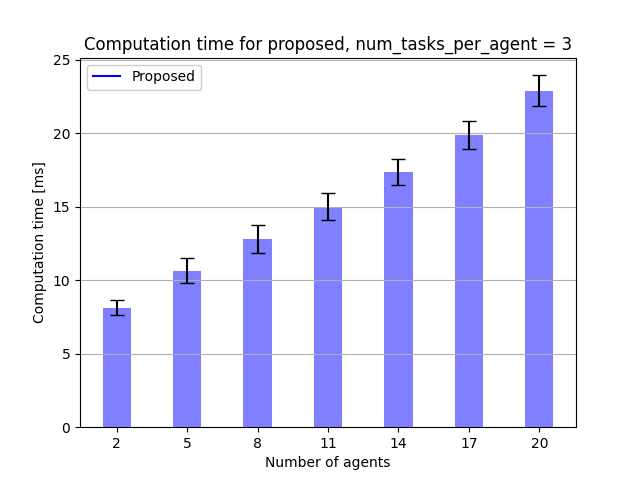}
		\caption{Computation time of DrMaMP}
		\label{comp:fixed_target:time_one}
	\end{subfigure}
	\hfill
	\begin{subfigure}{.30\textwidth}
		\centering
		\includegraphics[width=\linewidth]{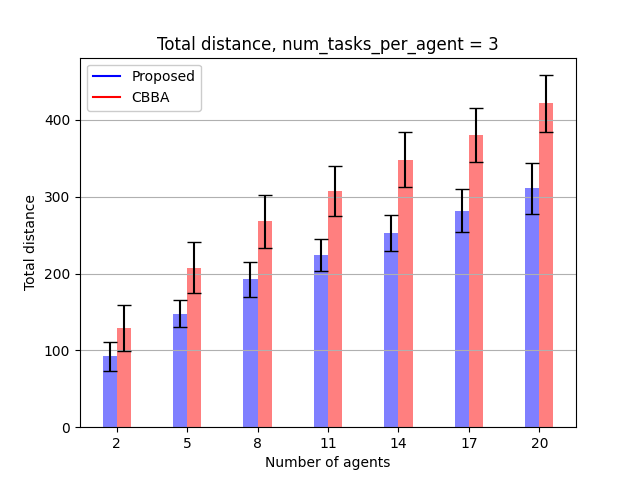}
		\caption{Total distance of two methods}
		\label{comp:fixed_target:distance_two}
	\end{subfigure}
	\caption{The computation time and total distance results of two methods: DrMaMP and \cite{bertuccelli2009real} with an increased number of agents. The number of unassigned tasks $n_t = 3n_a$. $N=300$.} 
	\label{comp:fixed_target}
\end{figure*}

In Fig. \ref{comp:fixed_target:time_two} and Fig. \ref{comp:fixed_target:time_one}, the computation time of \cite{bertuccelli2009real} is increased exponentially and is up to over 4.5 seconds when there are 20 agents and 60 tasks, whereas the computation time of DrMaMP is increased linearly, on the order of milliseconds.
The fully centralized method \cite{ren2021ms} has a similar scenario with a 32 $\times$ 32 map and random obstacles. According to Fig. 3 of \cite{ren2021ms}, it takes about 55 seconds to generate sequences for 5-20 agents and 10-50 tasks.
This paper omits the comparison with \cite{ren2021ms} because \cite{ren2021ms} is not a real-time algorithm.
The CBBA's computation time is increased exponentially because all agents need to take auctions iteratively and repeat for every task. The negotiation process for each task is more time-consuming and less efficient when $n_a$ is larger. Whereas for DrMaMP, the increased $n_a$ only raises the burden of Algorithm \ref{alg:algo_task_segment} and \ref{alg:cluster_assignment} slightly. The most computationally heavy part of DrMaMP is finding the collision-free path between every pair of nodes in each sub-problem, i.e., Line 4 - Line 7 of Algorithm \ref{alg:single}. Since Algorithm \ref{alg:single} is distributed over agents, the increased $n_a$ does not raise the computational load significantly. Section \ref{subsec:comput_analysis} analyzes the computational burden of DrMaMP and shows consistency with the comparisons.

As for optimality (total distance), DrMaMP outperforms \cite{bertuccelli2009real} because DrMaMP utilizes global information of tasks and agents in Algorithm \ref{alg:algo_task_segment} and \ref{alg:cluster_assignment}, while \cite{bertuccelli2009real} does auction for one task at one time. Thus the fully decentralized auction process does not utilize global information, resulting in less optimality. Moreover, the bid price in \cite{bertuccelli2009real} is the Euclidean distance between agent and task, and \cite{bertuccelli2009real} only generates collision-free paths after task order is determined. In a cluttered environment, the Euclidean distance is not the actual cost. Section \ref{subsec:comput_analysis} analyzes and compares the computational burden if \cite{bertuccelli2009real} utilizes collision-aware cost as the bid price.

\subsection{Comparison with Increased Number of Tasks}\label{subsec:comp_task}

\begin{figure*}[h]
	\centering
	\begin{subfigure}{.30\textwidth}
		\centering
		\includegraphics[width=\linewidth]{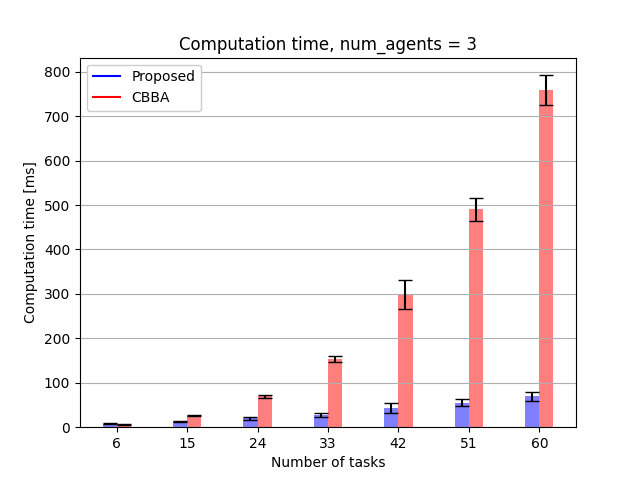}
		\caption{Computation time of two methods}
		\label{comp:fixed_agent:time_two}
	\end{subfigure}
	\hfill
	\begin{subfigure}{.30\textwidth}
		\centering
		\includegraphics[width=\linewidth]{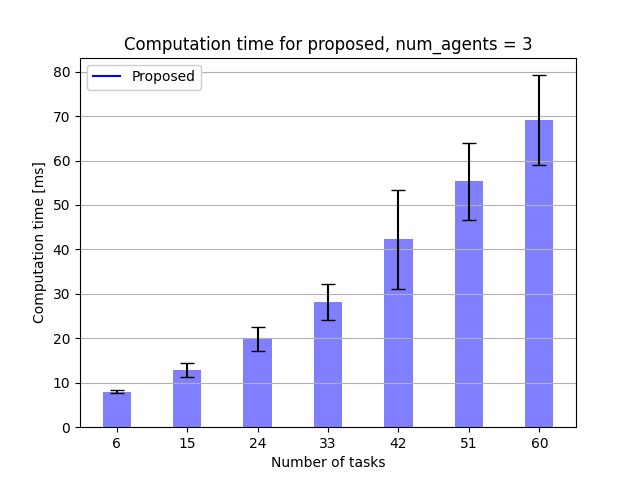}
		\caption{Computation time of DrMaMP}
		\label{comp:fixed_agent:time_one}
	\end{subfigure}
	\hfill
	\begin{subfigure}{.30\textwidth}
		\centering
		\includegraphics[width=\linewidth]{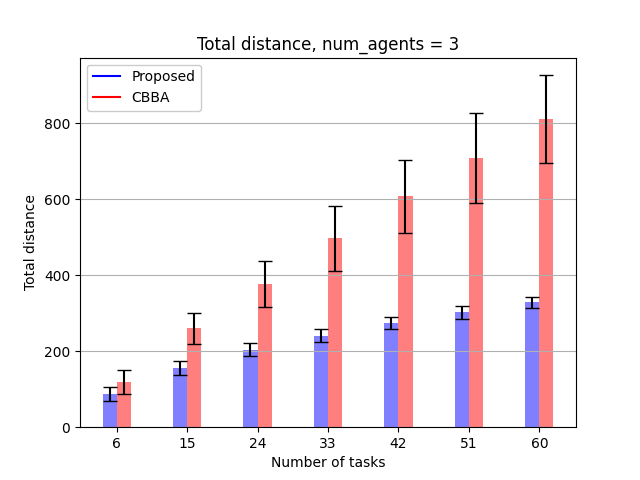}
		\caption{Total distance of two methods}
		\label{comp:fixed_agent:distance_two}
	\end{subfigure}
	\caption{The computation time and total distance results of two methods: DrMaMP and  \cite{bertuccelli2009real} with an increased number of tasks. The number of agents is fixed at 3. $N=300$.} 
	\label{comp:fixed_agent}
\end{figure*}

In Fig. \ref{comp:fixed_agent:time_two} and Fig. \ref{comp:fixed_agent:time_one}, CBBA's computation time is increased exponentially and is about 0.75 seconds for 60 tasks and 3 agents, whereas the computation time of DrMaMP is increased almost linearly. The increasing rate of CBBA's computation time in Fig. \ref{comp:fixed_agent:time_two} is much less than Fig. \ref{comp:fixed_target:time_two} because there is less negotiation among agents and thus the auction for each task needs fewer iterations when $n_a$ is smaller.

The increasing rate of DrMaMP's computation time in Fig. \ref{comp:fixed_agent:time_one} is greater than Fig. \ref{comp:fixed_target:time_one} because the linearly increased $n_t$ leads to the computational burden increasing quadratically (see Section \ref{subsec:comput_analysis}). Nevertheless, the magnitude of computation time is still relatively small because each agent only needs to deal with a task subset due to Algorithm \ref{alg:algo_task_segment}. The detailed analysis is shown in Section \ref{subsec:comput_analysis}. Fig. \ref{comp:fixed_agent:distance_two} shows that DrMaMP outperforms CBBA regarding optimality. These comparisons show that by using some global information in a distributed manner, DrMaMP achieves better performance than a decentralized method and a centralized method.



\subsection{Computational Burden Analysis}\label{subsec:comput_analysis}
DrMaMP approximates the traveling cost from one node to another one by the length of the underlying collision-free path. An intuitive way to improve the optimality of CBBA is to utilize the lengths of collision-free paths as bid prices. This subsection analyzes the computational burden of DrMaMP and this approach.

To find all possible paths, each agent connects to all the tasks and every two tasks connect to each other. Thus the total number of paths $\hat{N}_p$ for CBBA is
\begin{equation} \label{eq:num_path_cbba}
	\hat{N}_p = {}_{n_t}P_{2} + n_a \cdot n_t = n_t(n_t+n_a-1),
\end{equation}
where ${}_{n_t}P_{2}=\frac{n_t!}{(n_t-2)!}$ is the number of permutations for selecting two elements from total $n_t$ elements.

As for DrMaMP, the upper bound $\overline{N}_p$ for the number of paths to be found for each agent is $n_t$, i.e.,
\begin{equation}
	\overline{N}_p \triangleq \sup \max(|\hat{\mathcal{T}}_1|, \cdots, |\hat{\mathcal{T}}_{n_a}|) = n_t.
\end{equation}
Denote $\text{ceil}(\cdot): \mathbb{R} \mapsto \mathbb{Z}$ as the ceiling function, and $\text{ceil}(x)$ is the least integer greater than or equal to $x$. Since the entire task set is partitioned into $n_a$ subsets and the path-finding for each agent is parallel, the lower bound $\underline{N}_p$ is
\begin{equation}
	\underline{N}_p \triangleq \inf \max(|\hat{\mathcal{T}}_1|, \cdots, |\hat{\mathcal{T}}_{n_a}|) \triangleq n_c = \text{ceil}(\tfrac{n_t}{n_a}).
\end{equation}
Hence, the maximum number of paths $N_p$ in total for DrMaMP is
\begin{equation}
	n_c + {}_{n_c}P_{2} \leq N_p \leq n_t + {}_{n_t}P_{2} \ \Rightarrow \ {(\tfrac{n_t}{n_a})}^2 \lessapprox N_p \leq n_t^2.
\end{equation}
Combining with \eqref{eq:num_path_cbba} yields
\begin{equation}
	1 < 1 + \tfrac{n_a-1}{n_t} \leq \tfrac{\hat{N}_p}{N_p} \lessapprox (1+\tfrac{n_a-1}{n_t})n_a.
\end{equation}

Since ${(\frac{n_t}{n_a})}^2 \lessapprox N_p \leq n_t^2$, when $n_a$ is increased linearly and the ratio of $n_t$ to $n_a$ is a constant $a \triangleq \frac{n_t}{n_a}$, the lower bound of $N_p$ increases linearly as $\underline{N}_p = a^2 n_a$. This conclusion is consistent with Fig. \ref{comp:fixed_target:time_one}. Based on observation on comparisons, the actual computational burden of DrMaMP is skewed towards the lower bound. When $n_t$ is increased linearly and $n_a$ is fixed, $\underline{N}_p$ increases quadratically with respect to $n_t$. In addition, the standard deviation of computation time in Fig. \ref{comp:fixed_agent:time_one} is increasingly larger than in Fig. \ref{comp:fixed_target:time_one}. This observation happens because the number of assigned tasks for each agent $|\hat{\mathcal{T}}_1|, \cdots, |\hat{\mathcal{T}}_{n_a}|$ tends to be more diverse as $n_t$ increases and $n_a$ is constant. As for Fig. \ref{comp:fixed_target:time_one}, the task-agent ratio is fixed and thus the deviation remains relatively the same when $n_a$ increases.

As for replacing the bid cost as the length of a collision-free path, the extra computational burden of CBBA is greater than the actual burden of DrMaMP. The difference between the two upper bounds is $n_t(n_a-1)$, which increases linearly as $n_t$ or $n_a$ increases. When the task-agent ratio is fixed and $n_a$ increases, $\frac{\hat{N}_p}{\underline{N}_p}$ is still greater than 1 and it increases with a rate of $\frac{1}{n_a}$. The upper bound $\frac{\hat{N}_p}{\overline{N}_p}$ increases with a rate of $n_a$. When $n_a \ll n_t$, $ N_p \leq \hat{N}_p \lessapprox n_a \cdot N_p $. The worst case of DrMaMP is that its computation burden is slightly less than CBBA's, but CBBA's burden at most is $n_a$ times greater than DrMaMP's. Thus, task segmentation and parallelizable mission planning benefit run-time computation. And revising the bid prices of CBBA is not computationally efficient and hence the scalability is not good.


\subsection{Optimality Gap in Small-size Problems} \label{subsec:optimality_gap}

\begin{figure}[h]
	\centering
	\includegraphics[width=0.30\textwidth]{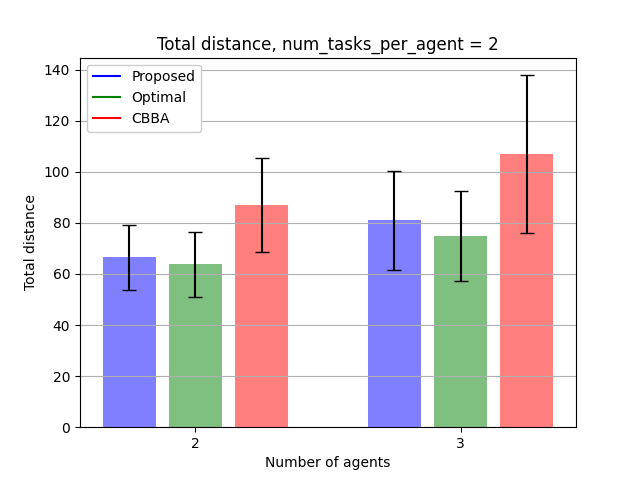}
	\caption{Optimality gap in small-size problems}
	\label{comp:optimality}
\end{figure}

Section \ref{subsec:optimality_gap} shows the optimality gap between DrMaMP and the global optimum. The global optimum is found by exhaustive search and thus the search is only feasible in small-size problems. Fig. \ref{comp:optimality} shows the optimality gap with two cases, 2 agents + 4 tasks and 3 agents + 6 tasks. For each case, there are 20 scenarios with different positions of agents, tasks, and obstacles. It is impossible to search a global optimum exhaustively for problems with a larger size since the MAMP problem is NP-hard.
The total number of solutions for $n_a$ agents and $n_t$ tasks is $\frac{(n_t+2n_a-1)!n_t!}{(n_a+n_t)!(n_a-1)!}$. 
For the case with 3 agents and 6 tasks, it has 39600 possible solutions and takes about 10 seconds to find a global optimum. For 4 agents and 8 tasks, it has 18345600 solutions and the estimated time to find an optimum is 78 minutes.
In the case of 2 agents, the optimality gap between DrMaMP and the optimum is on average 4.3\% while CBBA's cost is on average 36.3\% greater than the optimum. As for another case, DrMaMP's cost is on average 8.3\% greater than the optimum whereas CBBA's cost is 43.1\% greater than the optimum. The DrMaMP's optimality gap increases when the problem size increases since the task segmentation algorithm cannot explore all the permutations of the number of assigned tasks for each agent. Nevertheless, the algorithm makes the MAMP problem tractable and solves it at run-time.

\section{Conclusion}\label{sec:conclusion}
The collision-aware MAMP problem is NP-hard but requires real-time computational performance in many applications. This paper presents a distributed real-time algorithm DrMaMP. DrMaMP partitions the entire task set into several subsets such that each agent can determine the task allocation order and collision-free path parallelly. This process reduces the dimension of original problems and hence makes DrMaMP able to run in real-time with good scalability. The above results show that by using global information in a distributed manner, DrMaMP achieves better performance on both computation and optimality.


\bibliographystyle{ieeetr}
\bibliography{reference_this}

\end{document}